\documentclass{article}
\usepackage[preprint]{corl_2024} 

\usepackage{layouts}
\usepackage{subcaption}
\usepackage{amsmath}
\usepackage{algorithm}
\usepackage{booktabs}
\usepackage{wrapfig}
\usepackage{comment}

\usepackage{enumitem}

\usepackage[T1]{fontenc}   

\definecolor{fadedviolet}{rgb}{0.2,0.08,0.0.4}

\usepackage[beginLComment=/*~,endLComment=~*/]{algpseudocodex}
\algnewcommand{\LineComment}[1]{\State \(\triangleright\) \emph{\color{gray}#1}}
\algrenewcommand{\algorithmicrequire}{\textbf{Input:}}
\algrenewcommand{\algorithmicensure}{\textbf{Output:}}

\newcommand{\myparagraph}[1]{\textbf{#1}\quad{}}
\newcommand{\longoverbrace}[3]{\begingroup\color{gray}\overbrace{\color{black}#1}^{\centering\text{\parbox{#3}{\color{black}\hbox to 0cm{\hss #2 \hss}}}}\endgroup}
\newcommand{\longunderbrace}[3]{\begingroup\color{gray}\underbrace{\color{black}#1}_{\centering\text{\parbox{#3}{\color{black}\hbox to 0cm{\hss #2 \hss}}}}\endgroup}
\newcommand{\argmin}{\operatornamewithlimits{argmin}}
\newcommand{\apcost}{V_\text{A.P.}}

\newcommand{\OBJECTS}{\mathcal{O}}
\newcommand{\FLUENTS}{\mathcal{F}}
\newcommand{\ACTIONS}{\mathcal{A}}
\newcommand{\STATES}{\mathcal{S}}
\newcommand{\sgtuple}{\langle\sigma_g,k_g\rangle}
\newcommand{\kprep}{k_\text{\tiny{}prep}}

\usepackage[acronym]{glossaries}\usepackage{glossaries-extra}
\setabbreviationstyle[acronym]{long-short}
\setabbreviationstyle[short]{short-nolong}
\newacronym{TAMP}{\textsc{tamp}}{task and motion planning}
\newacronym{NAMO}{\textsc{namo}}{navigation among movable obstacles}
\newacronym{GNN}{\textsc{gnn}}{graph neural network}
\newacronym{PDDL}{\textsc{pddl}}{Planning Domain Definition Language}

\newcommand{\myopic}{\textsc{Myopic}}
\newcommand{\anttamp}{\textsc{AntTAMP}}
\newcommand{\prepmyopic}{\textsc{Prep+Myopic}}
\newcommand{\prepanttamp}{\textsc{Prep+AntTAMP}}
\glsdisablehyper{}

\usepackage{listings}
\def\BibTeX{{\rm B\kern-.05em{\sc i\kern-.025em b}\kern-.08em
    T\kern-.1667em\lower.7ex\hbox{E}\kern-.125emX}}
\usepackage{stfloats}
\usepackage{float}

\title{Anticipatory Task and Motion Planning}

\author{
Roshan Dhakal$^{\dagger}$,
Duc M. Nguyen$^\dagger$,
Tom Silver$^\mathsection$,
Xueso Xiao$^\dagger$,
Gregory J. Stein$^\dagger$\\
  $^\dagger$George Mason University\
  $^\mathsection$MIT Computer Science and Artificial Intelligence Laboratory \\ 
  \texttt{\{rdhakal2, mnguy21, xiao, gjstein\}@gmu.edu, tslvr@mit.edu}}

\begin{document}
\maketitle


\begin{abstract}
We consider a sequential task and motion planning (\textsc{tamp}) setting in which a robot is assigned continuous-space rearrangement-style tasks one-at-a-time in an environment that persists between each. Lacking advance knowledge of future tasks, existing (myopic) planning strategies unwittingly introduce side effects that impede completion of subsequent tasks: e.g., by blocking future access or manipulation. We present \emph{anticipatory task and motion planning}, in which estimates of expected future cost from a learned model inform selection of plans generated by a model-based \textsc{tamp} planner so as to avoid such side effects, choosing configurations of the environment that both complete the task and minimize overall cost. Simulated multi-task deployments in navigation-among-movable-obstacles and cabinet-loading domains yield improvements of 32.7\% and 16.7\% average per-task cost respectively. When given time in advance to \emph{prepare} the environment, our learning-augmented planning approach yields improvements of 83.1\% and 22.3\%. Both showcase the value of our approach. Finally, we also demonstrate anticipatory \textsc{tamp} on a real-world Fetch mobile manipulator. 

\end{abstract}



\section{Introduction}





We consider a sequential \gls{TAMP} setting, in which a long-lived robot is assigned rearrangement-style tasks one-at-a-time from a sequence. Notably, the environment persists between tasks, so that the terminal state after completing one task serves as the starting state for the next.
Lacking advance knowledge of what tasks the robot will later be assigned, existing \gls{TAMP} planners~\cite{garrett2020pddlstream, plaku2010sampling,toussaint2015logic,kaelbling2011hierarchical, srivastava2014combined,kim2019learning,chitnis2016guided,dantam2016incremental, khodeir_ral} are \emph{myopic}, targeting low-cost solutions to their immediate objective without regard to the future.
As such, this pervasive myopic planning strategy often incurs side effects on subsequent tasks that increase overall cost.

\begin{wrapfigure}{R}{0.65\textwidth}
    \vspace{-1.5em}
    \centering
    \includegraphics[width=0.62\textwidth]{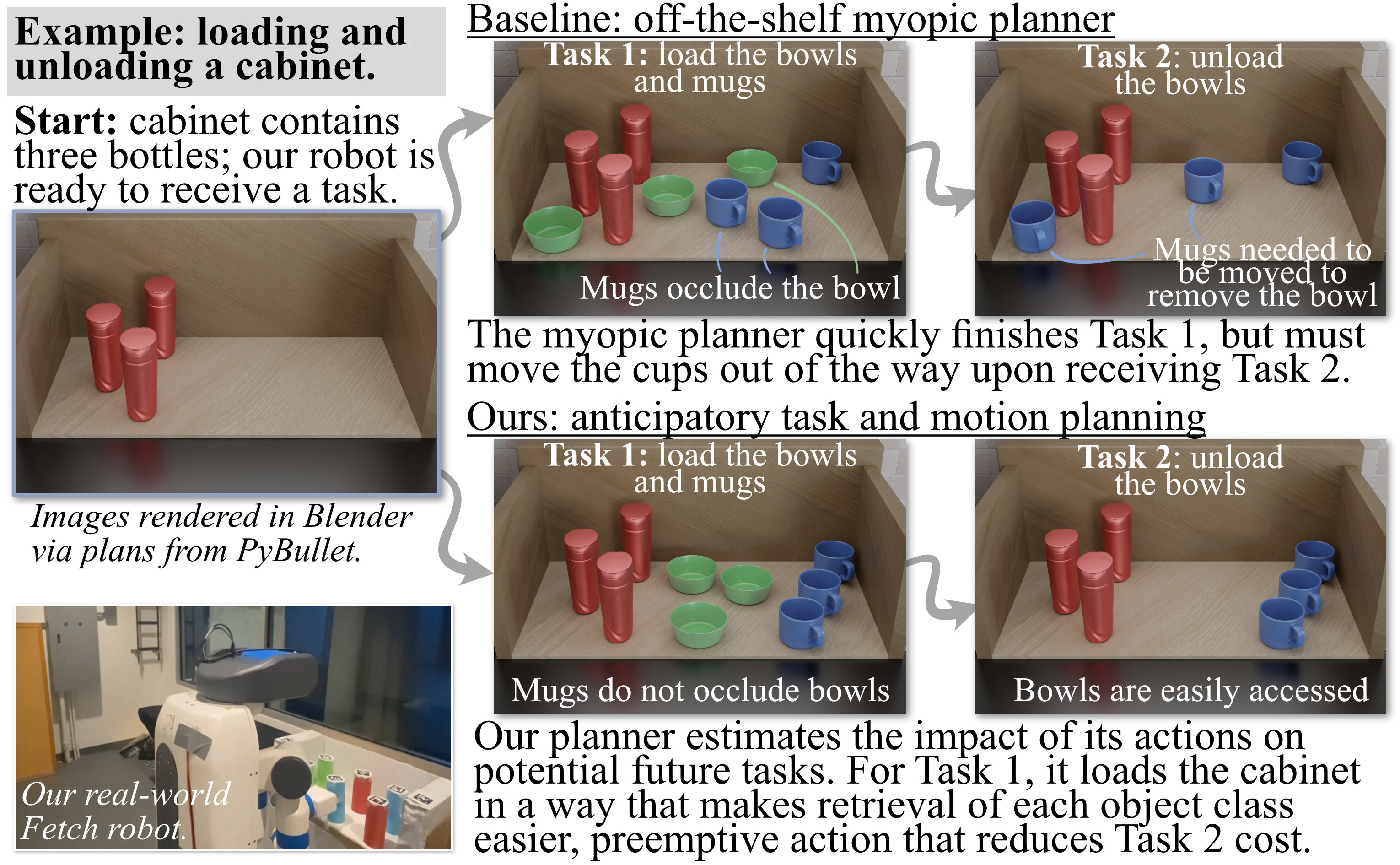}
    \vspace{-0.5em}
    \caption{Anticipatory \gls{TAMP} for a cabinet-loading scenario.}
    \label{fig:cabinet_example}
    \vspace{-1em}
\end{wrapfigure}

Consider the cabinet-loading scenario of Figure~\ref{fig:cabinet_example}. Myopic planning via off-the-shelf \gls{TAMP} solver~\cite{Chitnis2020CAMPsLC} quickly loads the mugs and bowls into the cabinet yet in a configuration that impedes completion of the second task to \texttt{unload the bowls}, for which the mugs must be moved out of the way. If the robot were to instead \emph{anticipate} that it may later be tasked to \texttt{unload the bowls} or \texttt{unload the mugs}, it would load the cabinet so as to avoid such side effects. As shown in Figure~\ref{fig:cabinet_example} (bottom): this small immediate expenditure of additional effort reduces overall cost.

The cabinet-loading scenario falls within the realm of \emph{anticipatory planning}~\citep{dhakal2023anticipatory}, an emerging subfield in which a robot jointly considers the cost of accomplishing its current task and the impact of its solution on subsequent tasks.
As it will not know its future tasks in advance, the robot must instead plan with respect to a \emph{task distribution}, which specifies what future tasks may later be assigned and their relative likelihood.
Anticipatory planning thus involves searching over the space of plans to find the one that jointly minimizes the immediate plan cost and the \emph{expected} future cost.



Recent work in the space of anticipatory planning~\cite{dhakal2023anticipatory, task-anticipation, patel2022proactive} so far focuses specifically on anticipatory \emph{task} planning problems, for which the state space is discrete.
However, rearrangement-style tasks in general often require jointly reasoning about both discrete elements of the state (e.g., whether an object is loaded into the cabinet) and continuous parameters of the state: e.g., where inside the cabinet the object is placed.
Integrated \glsentryfull{TAMP}, designed to solve such problems, is inherently complex due to the interconnected nature of the discrete and the continuous, since removing an object from the back of a cabinet may first require moving other objects out of the way. This challenge is further amplified by the need to anticipate how the robot's actions may negatively impact potential future tasks.
Existing anticipatory planning strategies are not well-suited to reason about these continuous aspects of the state, yet our cabinet-loading scenario illustrates the importance of their consideration: though loading all objects inside the cabinet specifies only a single \emph{symbolic state}, how those objects are arranged within the cabinet strongly determines how easily objects can be subsequently unloaded.
This work develops an approach that addresses this limitation of the state-of-the-art and so improves the performance of long-lived robots for which the environment persists between tasks.





We present \emph{Anticipatory Task and Motion Planning}, which improves the performance of long-lived robots over long deployments, consisting of sequences of tasks assigned one-at-a-time, by imbuing them with the ability to anticipate the impact of their actions on future tasks in continuously-valued manipulation and rearrangement tasks. Difficult to compute exactly, the expected future cost is estimated via learning using a \gls{GNN} that consumes a graphical representation of the state. A model-based \gls{TAMP} planner~\cite{srivastava2014combined, Chitnis2020CAMPsLC} generates candidate plans, in effect sampling over the continuous goal space, and we select the plan that minimizes the total cost: (i) the immediate cost produced by the \gls{TAMP} planner plus (ii) the estimated expected future cost from our learned estimator. Using learning and planning in tandem, our approach quickly and reliably completes its assigned objective while also producing solutions that reduce overall cost over lengthy deployments.

We evaluate the performance of our learning-augmented planning approach in two domains: an object-reaching scenario in a \gls{NAMO} domain and a cabinet-loading scenario. We demonstrate that our approach reduces average plan cost by 32.7\% over 20-task sequences in the \gls{NAMO} domain and by 16.7\% over 10-task sequences in the cabinet domain. Furthermore, if given time in advance to \emph{prepare} the environment before any tasks are assigned, we demonstrate performance improvements of 83.1\% and 22.3\% in the \gls{NAMO} and cabinet domains respectively. In both simulated and real-world experiments, we show the benefit of our learning-augmented \gls{TAMP} strategy, improving performance over deployments in persistent environments consisting of multiple tasks given in sequence, a step towards more performant long-lived robots.

\section{Related Work}

\myparagraph{Task and Motion Planning}
\Gls{TAMP} involves jointly reasoning over discrete (\emph{place the bowl}) and continuous (\emph{at pose $X$}) spaces to achieve long-horizon goals (\emph{load the bowls and mugs})~\cite{garrett2021integrated,zhao2024survey}. We build on the sampling-based \gls{TAMP} planner of~\citet{srivastava2014combined}. Existing \gls{TAMP} approaches~\cite{garrett2020pddlstream, plaku2010sampling, toussaint2015logic, kaelbling2011hierarchical, srivastava2014combined, dantam2016incremental, curtis2022discovering, bradley2022learning, khodeir_ral} solve tasks in isolation. We show empirically that this \emph{myopic} approach performs poorly when the environment persists and solutions to one task may impact the next.


\myparagraph{Anticipating and Avoiding Side Effects during Planning}
Recent research has similarly investigated how an understanding of the robot's actions on future tasks can improve planning, via estimation of expected future cost~\cite{dhakal2023anticipatory}, direct task prediction via \textsc{llm}s~\cite{task-anticipation}, or by leveraging human patterns of behavior~\cite{patel2022proactive}. 
These previous works consider discrete (task) planning, not integrated \gls{TAMP}.
On the contrary, we consider \gls{TAMP} problems in which planning requires both discrete and continuous elements of the state. Other work in the space of reinforcement learning~\cite{shah2019preferences} or in learning from demonstration~\cite{ravichandar2020lfdsurvey,manschitz2014learning,whitney2018learning,moro2018learning} seek to learn helpful behaviors by example or repeated interaction and so have potential for avoiding side effects, though are so far not directly applicable in our non-deterministic and long-horizon setting.




\myparagraph{Integrating Planning and Learning} To address some of the computational challenges in task planning and \gls{TAMP}, recent advances have leveraged learning: learning heuristics or using an \textsc{llm} for task-level planning~\citep{kim2019learning} or sampling distributions for continuous-space planning~\citep{kim2019learning, wang2018active,chitnis2019learning}.
However, all such approaches focus on solving a single task in isolation and are also not well-suited to address the unique challenges of our anticipatory \gls{TAMP} objective.
Our learned model for this work is a \glsentryfull{GNN}~\cite{battaglia2018relational, Shi2020MaskedLP, hamilton2017inductive}, as they have proven effective for \gls{TAMP} problems~\cite{silver2021planning, khodeir_ral, silver2021learning, kim2020learning}.

\section{Anticipatory Task and Motion Planning}\label{sec:antplan}
\subsection{Preliminaries}\label{sec:antplan:prelim}

\myparagraph{Task and Motion Planning}
We define a \glsentryfull{TAMP} problem as a tuple $\langle \OBJECTS, \FLUENTS, \ACTIONS, s_0, \tau \rangle$, following the convention of Chitnis et al.~\cite{chitnis2016guided}.
$\OBJECTS$ represents the set of objects, and so defines the configuration space $\STATES$ of the robot and all movable objects in the environment.
$\FLUENTS$ is the set of fluents, Boolean functions that describe the (discrete) symbolic relationships between objects.
$\ACTIONS$ is the set of high-level actions that the robot can execute to transform the state, such as \texttt{pick}, \texttt{move}, and \texttt{place}.
$s_0 \in \STATES$ is the initial state of the environment.
The task $\tau$ is defined as a set of fluents that define the goal state; as such, a task $\tau$ defines a subset of the state space---the goal region $G_\tau \subset \mathcal{S}$---in which the task is considered complete.

For notational convenience later on, we represent a state $s$ as a tuple $\langle \sigma, k \rangle$, representing the \emph{discrete symbolic components of the state} $\sigma$ (e.g., on which surface an object is placed) and the \emph{continuous aspects of the state} $k$: e.g., where on that surface the object is placed. 
The goal of a \gls{TAMP} solver is to find a sequence of actions that completes the specified task $\tau$, reaching a goal state $s_g \in G_\tau$.
In this work, we assume access to a \gls{TAMP} planner~\cite{srivastava2014combined, Chitnis2020CAMPsLC} that returns a plan $\pi$, a sequence of actions $a_0, a_1, \cdots, a_n \in \ACTIONS$, such that the terminal state of the plan is in the goal: $s_g \equiv \text{tail}(\pi) \in G_\tau$. 


\myparagraph{Anticipatory Planning}
There is often considerable flexibility in how the robot can choose to complete its assigned task $\tau$: i.e., the goal region $G_\tau$ consists of more than a single satisfying state.
During long-lived deployments, the environment will persist \emph{between tasks}, and so effective planning requires that the robot consider how its choice of how to complete the current objective---which goal state $s_g$ it ends up in---impacts possible tasks it may later be assigned, where tasks are assigned according to a \emph{task distribution} $P(\tau)$.
We adopt the formalism of \citet{dhakal2023anticipatory} that defines anticipatory planning as a joint minimization over (i) the cost to complete its assigned current task $\tau_c$ and (ii) the expected future cost to complete a subsequent task:\\[-4pt]
\begin{equation}\label{eq:antplan}
s^*_g
= \argmin_{s_g \in G(\tau_c)} \Big[ \longoverbrace{V_{s_g}(s_0)}{\parbox{12em}{\textbf{Immediate Task Cost:}\\[-3pt]Cost to reach $s_g$ from $s_0$}\hspace{5em}}{20em} + \longoverbrace{\sum_{\tau} P(\tau) V_{\tau}(s_g)}{\hspace{12em}\parbox{20em}{\textbf{Anticipatory Planning Cost:}\\[-3pt]Expected cost over future tasks}}{30em} \Big]
~,
\end{equation}\\[-4pt]
where $V_{s_g}(s_0)$ is the plan cost to reach state $s_g$ from starting state $s_0$ and $V_\tau(s_g)$ is the cost to complete task $\tau$ from state $s_g$.
As mentioned by~\citet{dhakal2023anticipatory}, reasoning about expected future cost over long sequences of tasks is computationally challenging, and so Eq.~\eqref{eq:antplan} instead seeks to minimize cost over an immediate task and a single next task in the sequence; we will show in Sec.~\ref{sec:exp} that this formulation is still sufficient for improved behavior over lengthy sequences.

Owing to the difficulty of integrated \gls{TAMP}, recent work in this space~\cite{dhakal2023anticipatory, task-anticipation, patel2022proactive} considers only \emph{task planning} settings, focusing only on symbolic planning and thus ignoring continuously-valued aspects of the state, a restriction we overcome in this work.

\subsection{Problem Formulation: Anticipatory Task and Motion Planning}\label{sec:antplan:prob}

We solve the problem of anticipatory \gls{TAMP}, which combines elements of both \gls{TAMP} and anticipatory planning and so is defined by the tuple:
$\langle \OBJECTS, \FLUENTS, \ACTIONS, s_0, \tau, P(\tau) \rangle$.
Tackling this problem requires reasoning about both discrete and continuous aspects of the state and so anticipatory \gls{TAMP} in general involves solving the following objective:\\[-4pt]
\begin{equation}\label{eq:cont_antplan}
\sigma^*_g, k^*_g
\! = \!\!\!\!\argmin_{\sigma_g,k_g \in G(\tau_c)}\!\! \Big[ \longoverbrace{V_{\sigma_g, k_g}\!(s_0)}{\parbox{12em}{\textbf{Immediate Task Cost:}\\[-3pt]Cost to reach $s_g \equiv \sgtuple$}\hspace{5em}}{20em} \!+\! \longoverbrace{\sum_{\tau} P(\tau) V_{\tau}(\sgtuple)}{\hspace{12em}\parbox{20em}{\textbf{Anticipatory Planning Cost:}\\[-3pt]Expected cost over future tasks}}{30em} \Big]
\! = \!\!\!\!\argmin_{\sigma_g,k_g \in G(\tau_c)}\!\! \Big[ V_{\sigma_g, k_g}\!(s_0) + \longoverbrace{\apcost{}(\sgtuple)}{\hspace{12em}\parbox{20em}{$\apcost{}$ is shorthand for\\[-3pt]expected future cost.}}{20em} \Big]~,
\end{equation}
where $V_{\sigma, k}(s_0)$ is the plan cost to reach state $s \equiv \langle \sigma, k \rangle$ from initial state $s_0$ and $V_\tau(\sgtuple)$ is the cost to complete task $\tau$ from state $s_g \equiv \sgtuple$.

\myparagraph{Preparation as task-free anticipatory \gls{TAMP}}
Household robots will not be in perpetual use and so can take preemptive action before any tasks are assigned to \emph{prepare} the environment: rearranging their environments to reduce expected future costs and thus make it easier to complete tasks once they are eventually assigned.
Formally, preparation can be defined as \emph{task-free anticipatory planning}~\cite{dhakal2023anticipatory}. For anticipatory \gls{TAMP}, we define preparation as minimization over the space of continuous states $k$ associated with the current symbolic state $\sigma_0$: $\kprep \in K(\sigma_0)$.
Thus, the prepared state of the environment $\langle\sigma_0,\kprep^*\rangle$ is defined via:
\begin{equation}\label{eq:prep}
\kprep^*
= \argmin_{\kprep \in K(\sigma_0)} \big[ \apcost{}(\langle \sigma_0, \kprep \rangle) \big]
\end{equation}


\subsection{Approach: Planning via Anticipatory \textsc{Tamp}}
\label{sec:anttampalgorithm}

\begin{wrapfigure}{R}{0.55\textwidth}
    \vspace{-1em}
    \centering
    \includegraphics[width=0.55\textwidth]{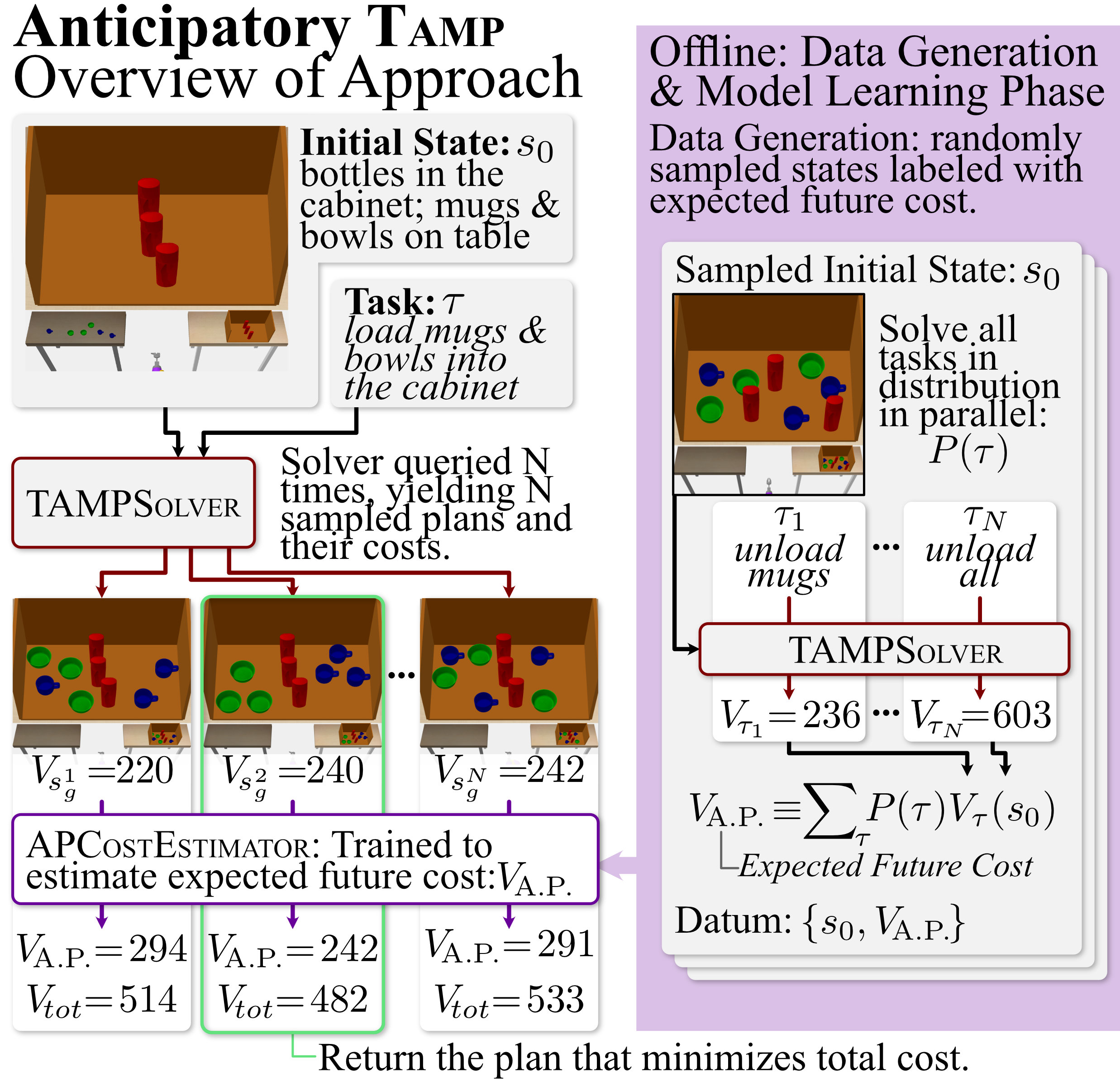}
    \caption{\textbf{Schematic of our approach}.}\label{fig:flowchart}
    \vspace{0.5em}
\end{wrapfigure}

Anticipatory task and motion planning requires that we search over plans (and thus the continuous goal space) and select the plan that minimizes the total cost, the sum of the immediate plan cost and the expected future cost, the anticipatory planning cost $\apcost{}$, as in Eq.~\eqref{eq:cont_antplan}. Difficult to compute at planning time, we rely on a learned model to estimate the $\apcost{}$ during planning: $\textsc{APCostEstimator}$. Figure~\ref{fig:flowchart} and Algorithm~\ref{alg:anttamp} each give an overview of planning via anticipatory \gls{TAMP} from an initial state $s_0$ and task $\tau$.



\begin{wrapfigure}{R}{0.48\textwidth}
\vspace{-1.5em}
\begin{minipage}{0.48\textwidth}
\begin{algorithm}[H]
\caption{Anticipatory \textsc{Tamp}}\label{alg:anttamp}
\algrenewcommand{\algorithmiccomment}[1]{\hfill\textbf{//}\,#1}
\scriptsize
\begin{algorithmic}[0]
\Function{AnticipatoryTAMP}{$s_0, \tau, \textsc{APCostEstimator}$}
\State $V^*_\text{total}= \infty$
\For{$i \in \{1, 2, \dots, N\}$}
    \LineComment{Get a candidate plan and its cost for $\tau$ using off-the-shelf TAMP solver, thus sampling goal states.}
    \State $\pi, \textsc{PlanCost} = \textsc{TAMPSolver}(s_0, \tau)$
    \LineComment{Get the final state for the candidate plan.}
    \State $\langle\sigma_g, k'_g\rangle = \textsc{Tail}(\pi)$
    \LineComment{Myopic planning uses a zero-function in place of the $\apcost{}$ estimator, as it does not anticipate future tasks.}
    \State $\apcost{} = \textsc{APCostEstimator}(\langle\sigma_g, k'_g\rangle)$
    \State $V_\text{total} = \textsc{PlanCost} + \apcost{}$
    \If{$V_\text{total}\leq V^*_\text{total}$}
        \State $\pi^* = \pi$
        \State $V^*_\text{total}(\langle\sigma_g, k^*_g\rangle) = V_\text{total}$
    \EndIf
\EndFor
    \State \Return $\pi^*$
\EndFunction
\end{algorithmic}
\end{algorithm}
\end{minipage}
\vspace{-1em}
\end{wrapfigure}

Our algorithm relies on an off-the-shelf \gls{TAMP} solver~\cite{Chitnis2020CAMPsLC, srivastava2014combined}.
\textsc{TAMPSolver} is a randomized planner that produces multiple candidate plans with varying continuous goal states (see the Appendix for details).
As such, each call to the planner returns plan $\pi$ that satisfies the task and terminates in a random state within the goal region. This property facilitates its use as a random sampler of goal states.
Under our anticipatory \gls{TAMP} approach, we query $\textsc{TAMPSolver}$ $N$ times, yielding $N$ plans and the plan cost of each. For each plan, we estimate $\apcost{}$ via our learned model $\textsc{APCostEstimator}$ and return the plan that minimizes total cost according to Eq.~\eqref{eq:cont_antplan}.

In general, search over both discrete and continuous aspects of the state is computationally demanding. As such, this work specifically considers relatively \emph{well-specified} tasks, in which the goal region specified by the task corresponds only a limited number of symbolic states, so that search for anticipatory \gls{TAMP} emphasizes the continuous aspects of the state: e.g., the goal region for the task \emph{load all objects into the cabinet} consists of only a single symbolic goal state $\sigma_g$ in which all objects must be inside the cabinet, yet their continuous pose within the cabinet is unspecified.

\emph{Preparation}, task-free anticipatory \gls{TAMP}, involves searching for the continuously-valued state that minimizes expected future cost in advance of being given a task.
We perform this search via a simulated annealing optimization approach~\cite{sim_annealing}. 
Beginning with an initial state $\langle\sigma_0, k_0\rangle$, we iteratively perturb the continuous object states $k$ within a bounded range ensuring no overlaps. If the $\apcost{}$ of the new state, estimated by $\textsc{APCostEstimator}$, is improved or meets a probabilistic criterion influenced by a decreasing temperature factor, it is accepted as the new state to perturb.
Search proceeds for $N$ iterations, eventually returning the prepared state $\langle\sigma_0, k^*_\text{\tiny{prep}}\rangle$.





\section{Estimating Expected Future Cost via Graph Neural Networks}\label{sec:learning}

During deployment, direct computation of the anticipatory planning cost $\apcost{}(\langle\sigma_g, k_g\rangle)$ is typically infeasible, either due to the high computational cost of solving all possible future tasks for every state considered during planning or the lack of direct access to the underlying task distribution $P(\tau)$.
Instead, we estimate $\apcost{}$ via an estimator (\textsc{APCostEstimator}), a \gls{GNN}~\cite{battaglia2018relational, Shi2020MaskedLP} trained via supervised learning with data generated during from an offline training phase.

\myparagraph{Training Data Generation}\label{sec:learning:data}
Offline, we presume that the robot has direct access to the underlying task distribution $P(\tau)$, which specifies what tasks the robot may be assigned and their relative likelihood.
We generate training data by randomly sampling states from the domain of interest.
For each state $s_i$, we solve every possible future task $\tau$ using \textsc{TAMPSolver} and use the resulting plan costs $V_\tau(s_i)$ to compute the anticipatory planning cost via its definition: $\apcost{} \equiv \sum_{\tau}P(\tau)V_\tau(s_i)$.
Each datum consists of states $s_i$ with labels $\apcost{}(s_i)$, which the learned model is then trained to estimate.

\begin{wrapfigure}{R}{0.5\textwidth}
    \vspace{-0.5em}
    \includegraphics[width=0.48\textwidth]{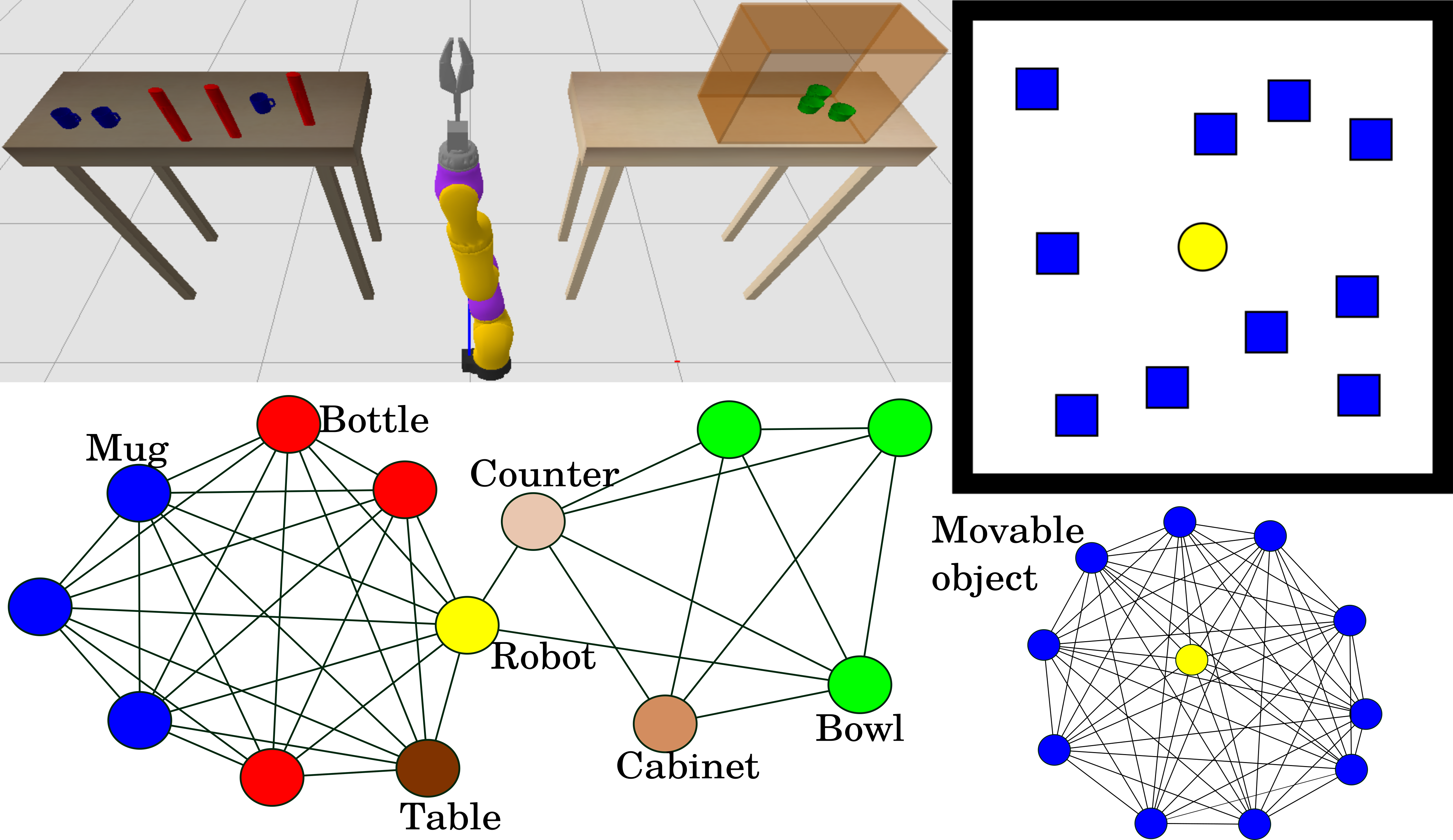}
    \caption{Example states and visualizations of their accompanying graph representations for both our Cabinet and \gls{NAMO} domains.}
    \label{fig:stategraph}
    \vspace{-0.5em}
\end{wrapfigure}

\myparagraph{Learning and Estimation via Graph Neural Networks}
As we rely on a \gls{GNN} for learning and estimation, we represent the environment state $\langle\sigma, k\rangle$ as a graph $\mathcal{G}$, as shown in Figure~\ref{fig:stategraph}. Nodes represent objects---e.g., the robot, movable objects, and object containers---and edges represent spatial or semantic relationships between them.
$\mathcal{G}$ includes features for both nodes and edges, specific to each environment; see details alongside our experiments in Sec.~\ref{sec:exp}.

Our \gls{GNN} is implemented via PyTorch Geometric~\cite{pyg} and consists of three TransformerConv layers~\cite{Shi2020MaskedLP} each followed by a leaky ReLU activation, culminating in a mean-sum pooling operation and a fully-connected layer. As our target is a regression objective, we use a mean absolute error loss and train with AdaGrad~\cite{duchi11a} with a batch size of 8 for 10 epochs with a 0.05 learning rate.

\section{Experiments and Results}\label{sec:exp}
\begin{figure}[t]
    \includegraphics[width=\textwidth]{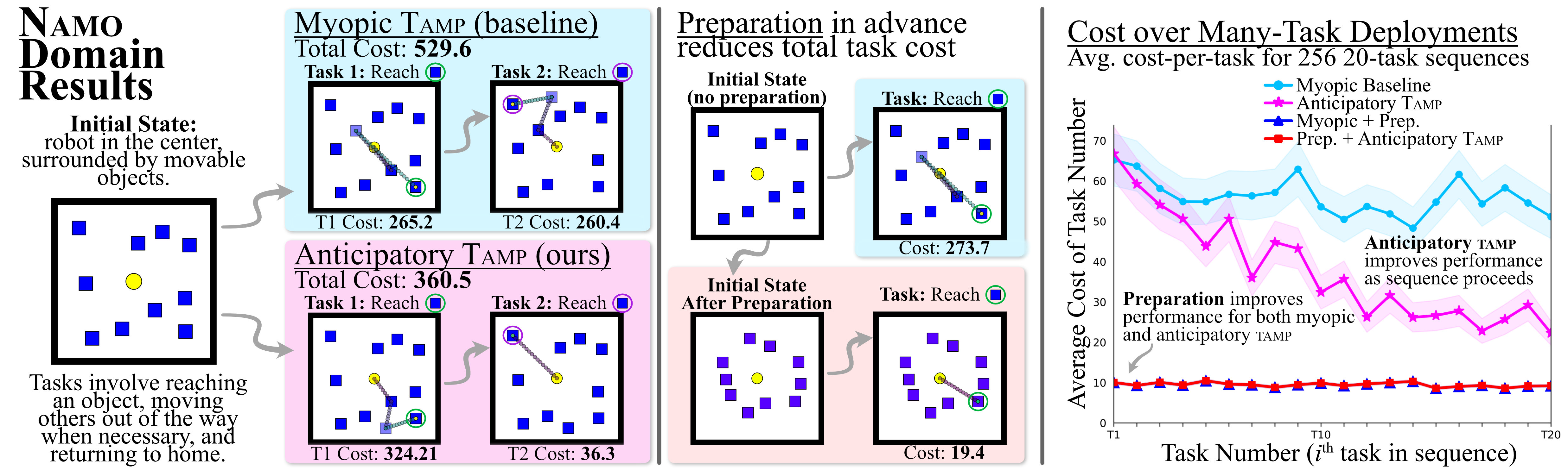}
    \caption{Results from our \glsentryfull{NAMO} domain. In examples (left), \anttamp{} moves the obstacle so that it is out of the way for future tasks, unlike the \myopic{} plan, reducing cost for subsequent tasks. Visualized plans omit the robot's return-to--home for clarity. 
    Right, we plot the average per-task performance over time (i.e., as each sequence proceeds), demonstrating that our anticipatory \gls{TAMP} tends towards reducing average task cost over its deployment.}\label{fig:namo_results}
\end{figure}


We evaluate our approach on two PyBullet~\cite{coumans2021} simulated domains based on those by \citet{Chitnis2020CAMPsLC}: (1) cabinet-loading and (2) object-reaching in a navigation among movable obstacles (\textsc{namo}) domain. We include trials for planning both with and without our anticipatory \gls{TAMP} approach, starting from either (i) a randomized initial state or (ii) a \emph{prepared} state (Sec.~\ref{sec:antplan:prob}). For each trial, we evaluate performance of four planners:

\begin{LaTeXdescription}
  \item[\myopic] Myopic \gls{TAMP}, which does not anticipate future tasks. Planning relies on Alg.~\ref{alg:anttamp} using a zero-function for the anticipatory cost estimator, ensuring fair comparison with our approach.
  \item[\anttamp] Our anticipatory \gls{TAMP} approach, which seeks to minimize both immediate and expected future cost via Eq.~\eqref{eq:cont_antplan}. We plan via Alg.~\ref{alg:anttamp} using a learned anticipatory cost estimator, trained in the environment of interest.
  \item[\prepmyopic]  The robot first \emph{prepares} the scene via Eq.~\eqref{eq:prep} and then plans via \myopic{}.
  \item[\prepanttamp] The robot first \emph{prepares} via Eq.~\eqref{eq:prep} and then plans via \anttamp{}.
\end{LaTeXdescription}

\subsection{Object Reaching in a Navigation Among Movable Obstacles (NAMO) Domain} \label{sec:exp:namo}
\begin{wraptable}{R}{0.45\textwidth}
    \vspace{-1em}
    \small
    \centering
    \begin{tabular}{cc}
    \toprule
    Planner & Average Cost \\
    \midrule
    \myopic{} & 56.1 \\
    \anttamp{} (ours) & \textbf{37.8} \\
    \midrule
    \textsc{Prep} (ours)+\textsc{Myopic} & \textbf{9.5} \\
    \prepanttamp{} (ours) & \textbf{9.5} \\ 
    \bottomrule
    \end{tabular}
    \caption{Average cost-per-task over 20-task sequences in our \gls{NAMO} domain.}
    \label{table:namo}
    \vspace{-10pt}
\end{wraptable}

We first perform experiments in a \glsentryfull{NAMO} domain, in which the robot must navigate to a target object, ostensibly so that the object may be interacted with or inspected, and then return to its starting position. The robot's task distribution is uniform, so that each of the 10 objects is chosen with uniform probability.
Reaching the target object may require moving other objects out of the way, and so good performance in this environment in general will require that the robot position objects so that they do not block the path to others.
As discussed in Sec.~\ref{sec:learning}, our anticipatory cost estimator is a \gls{GNN}, trained with 10\textsc{k} data. Each node corresponds to an entities in the scene (e.g., the robot and movable objects) with input features of a one-hot vector of the entity class, the object pose, and its distance to the robot. Edge features include distances between each node and the number of movable obstacles between them.
Evaluation consists of 256 trials, each a random sequence of 20 tasks; 
planning via Algorithm~\ref{alg:anttamp} uses 100 samples of candidate plans per task and 5000 samples per preparation.


Table~\ref{table:namo} shows the average cost-per-task for all four planning strategies and thus the benefits of using our approaches, both \anttamp{} and preparation. \anttamp{} has a 32.7\% lower overall planning cost compared to \myopic{}.
Moreover, advance preparation of the environment further reduces the cost of both strategies---an 83.1\% improvement over \myopic{} alone. 
Notably, performance for \prepmyopic{} and \prepanttamp{} is \emph{identical}: preparation routinely finds states from which all objects can be reached without needing to move any others out of the way, so that all tasks are quickly and easily completed; thus, anticipatory \gls{TAMP} does not go out of its way to rearrange the blocks during task execution, as doing so would increase overall cost.

Figure~\ref{fig:namo_results} (right) shows the average per-task performance for each strategy: i.e., a downward slope indicates that the cost of the final task averaged across the 256 sequences is less than the average cost of the first task. In particular, the average cost of planning via \anttamp{} decreases over time, showing how our approach gradually makes the environment easier to use through repeated interaction, an emergent property of planning via anticipatory \gls{TAMP}.
We highlight a few examples in Figure~\ref{fig:namo_results} that corroborate the quantitative results, namely that planning via \anttamp{} ensures that obstacles are placed so as to be out of the way for subsequent tasks.
In particular, the prepared states are more \emph{ring-like}, so that the robot can reach all objects without moving any out of the way.

\subsection{Cabinet Loading and Unloading Scenario} \label{sec:exp:cabinet}

\begin{wraptable}{R}{0.45\textwidth}
    \vspace{-1em}
    \small
    \centering
    \begin{tabular}{cc}
    \toprule
    Planner & Average Cost \\
    \midrule
    \myopic{} & 283.2 \\
    \anttamp{} (ours) & \textbf{235.8} \\
    \midrule
    \textsc{Prep} (ours)+\textsc{Myopic} & 267.7 \\
    \prepanttamp{} (ours) & \textbf{219.9} \\
    \bottomrule
    \end{tabular}
    \caption{Average cost-per-task over 20-task sequences in our Cabinet domain.}
    \label{table:cabinet}
    \vspace{-10pt}
\end{wraptable}

Our cabinet domain consists of nine objects: three each of mugs (blue), bottles (red), and bowls (green), using \textsc{urdf} models from~\cite{liu2021ocrtoc}.
In this domain, the robot's tasks involve loading or unloading all objects belonging to one or more semantic class---e.g., move all bottles to the table---each assigned with equal probability. 
To estimate the anticipatory planning cost, we train a \gls{GNN} with 5\textsc{k} data.
Node features consist of a one-hot encoding of entity type (robot, container, or object), pose, and distance from the robot. Edges represent spatial or semantic relationships between entities and edge features include the distances between nodes they connect and the number of movable obstacles between them, with zero used for non-movable objects. Due to the interchangeable nature of objects of the same semantic class, only objects of other semantic classes are regarded as obstacles for the purposes of edge feature computation. 
Evaluation consists of 64 deployments, each a sequence of 10 tasks; planning via Algorithm~\ref{alg:anttamp} internally uses 200 samples of candidate plans per task and 2500 samples per preparation.

\begin{figure}[t]
    \centering
    \includegraphics[width=\textwidth]{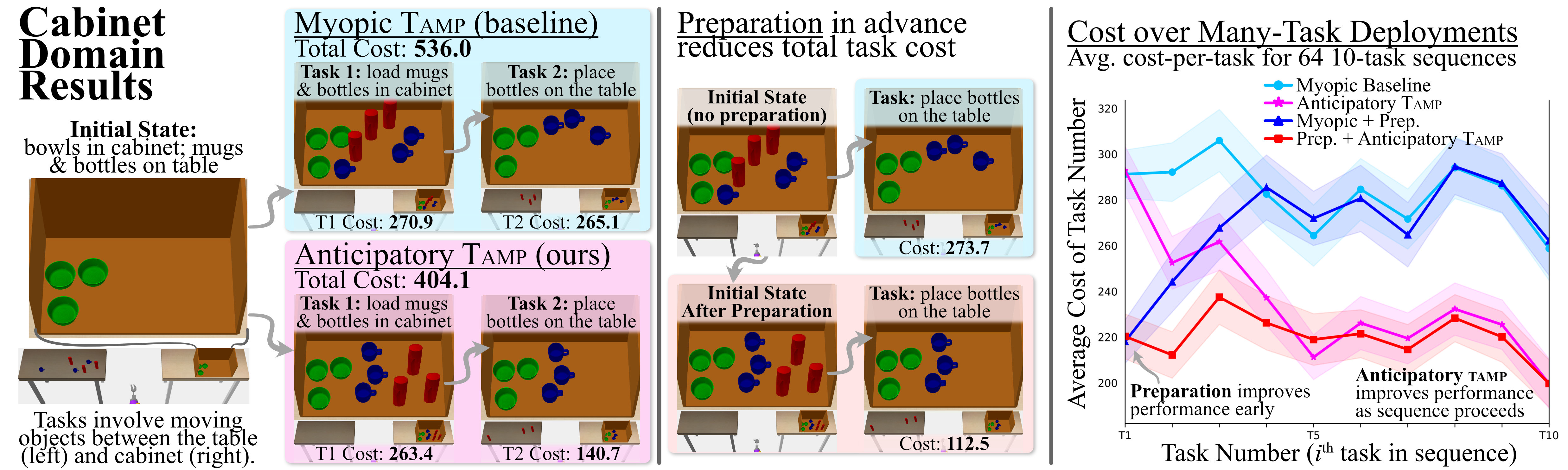}    
    \caption{Results from our cabinet domain. In examples (left), \anttamp{} anticipates future tasks, reducing total planning cost compared to myopic planning. Right, we plot the average per-task performance over time (i.e., as each sequence proceeds), demonstrating that our anticipatory \gls{TAMP} tends towards reducing average task cost over its deployment.}\label{fig:cabinet_results}
    \vspace{-1em}
\end{figure}


Table~\ref{table:cabinet} shows the average cost-per-task for all four planning strategies; our approach yields performance benefits for both anticipatory \gls{TAMP} and preparation. The cost-over-time plot in Figure~\ref{fig:cabinet_results} (right) shows how \anttamp{} from a random initial state yields performance improvements over \myopic{} that increase over deployment: our approach makes the environment gradually easier to use over time. Preparation benefits both approaches, yet for \myopic{} those benefits diminish over time, owing to its lack of consideration for tasks the robot may be asked to perform later on. Combining \prepanttamp{} yields the best of both strategies.
Figure~\ref{fig:cabinet_results} shows example trials, which support our statistical results; each shows how planners that rely on our anticipatory \gls{TAMP} have discovered the benefit of loading objects so as to avoid obstructing those of different semantic class and so tend to reduce overall cost over the course of each 10-task sequence.

\subsection{Real-World Demonstration on a Fetch Mobile Manipulator}\label{sec:real_robot}

\begin{wrapfigure}{R}{0.6\textwidth}
    \vspace{-2em}
    \includegraphics[width=0.6\textwidth]{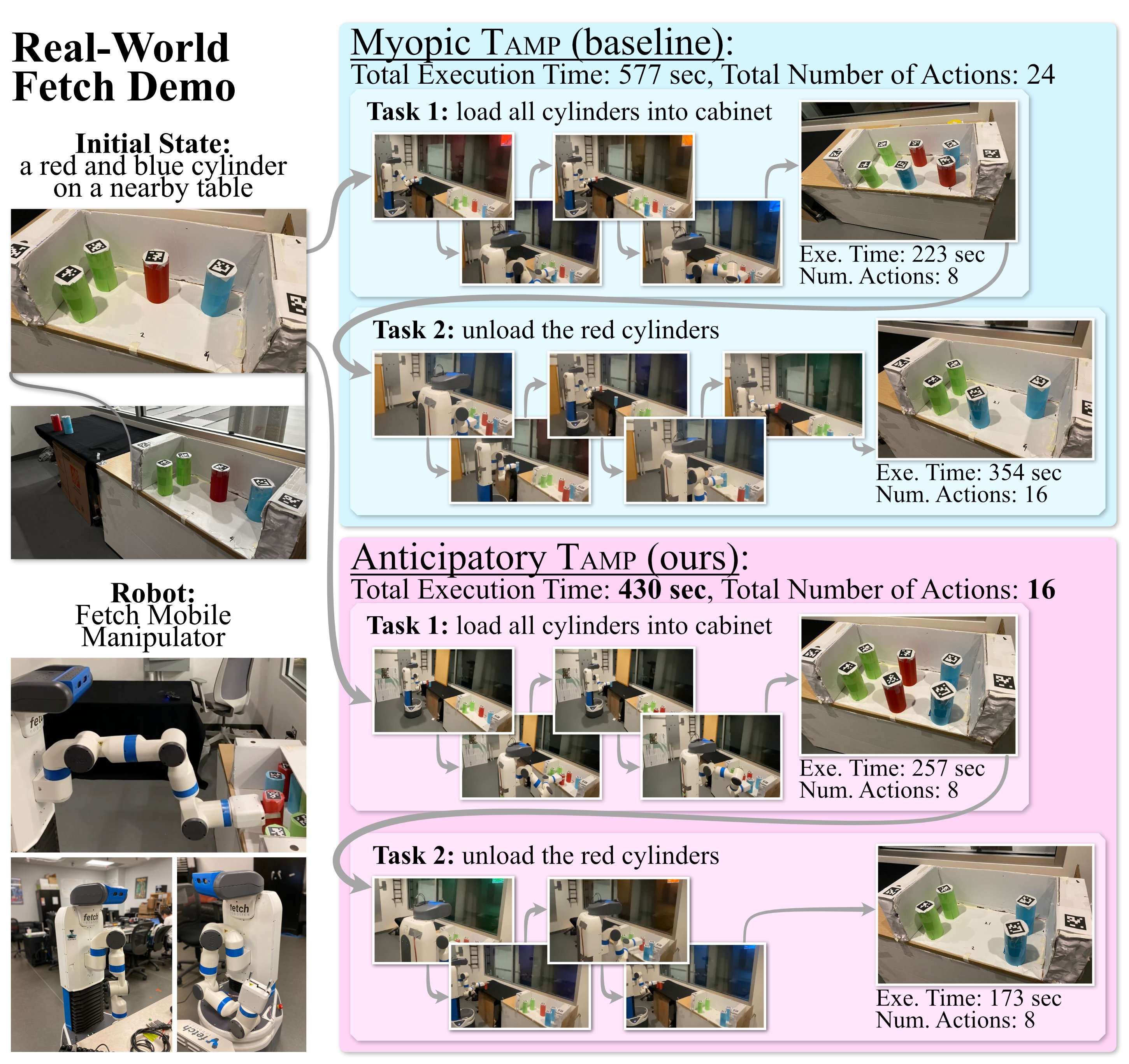}    
    \vspace{-1.4em}
    \caption{Real-world demonstration with the Fetch.}
    \label{fig:fetch}
    \vspace{-1em}
\end{wrapfigure}


We further evaluate our approach using the Fetch Mobile Manipulator~\cite{fetch2023} in a real-world cabinet-loading scenario with six cylinders---two each in red, blue, and green---and with an anticipatory cost estimator trained in simulation environments of the same composition.
We show the performance of both \anttamp{} and \myopic{} planners; both are initialized with the same starting configuration (Figure~\ref{fig:fetch} left) and tasked to load the two objects from the table into the cabinet. Upon completion of that first task, each planner is then instructed to unload the two red cylinders and move them to the table, a task made difficult if either red cylinder is obstructed by another non-red cylinder.

The \myopic{} planner (Figure~\ref{fig:fetch} top), incapable of considering the effect of its actions on possible future tasks, solves the first task such that the back red cylinder is more difficult to remove when the second task is eventually assigned.
Conversely, our \anttamp{} approach places the cylinders in same-colored groups and so the second task is made easier.
Our approach requires fewer total actions (pick, move, place) and correspondingly less total execution time to complete the tasks in the sequence, as illustrated in Figure~\ref{fig:fetch}.

\section{Discussion, Limitations, and Future Work}
\label{sec:conclusion}

We present anticipatory \glsentryfull{TAMP}, a planning strategy to improve performance over task sequences in persistent continuous-space rearrangement-style settings.
Unlike most existing \gls{TAMP} approaches, which focus only on one task at a time, we train a learned model to estimate how the robot's immediate actions will impact potential future tasks and use this \emph{anticipatory planning cost} to select plans that jointly minimize immediate plan cost and expected future cost. In both simulated and real world experiments, we show the benefit of our learning-augmented \gls{TAMP} strategy, improving performance over deployments in persistent environments consisting of multiple tasks given in sequence, an important step towards more performant long-lived robots.


\textbf{Limitations}\quad{}
Our approach, which relies on existing solvers during planning, suffers from many of the same challenges of scale and computation that limit \gls{TAMP} for rearrangement problems in general.
Moreover, planning via Algorithm~\ref{alg:anttamp} requires running the underlying \gls{TAMP} solver many times (on the order of 100 in our experiments) to effectively search the continuous goal space. As such, better sampling heuristics or improvements in search will likely be necessary to scale up to more complex problems, in part why our work so far has focused on tasks for which the final symbolic state is presumed relatively well-specified.
Finally, we presume offline access to the underlying task distribution, which defines our objective function. Not only may the robot not have this access in advance, but we also so far presume that the online task distribution matches that seen during training and will not evolve during deployment; future work may consider learning this task distribution online and modeling its change over time.


\acknowledgments{This material is based upon work supported by the National Science Foundation under Grant No. 2232733.}



\bibliography{references}  

\begin{thebibliography}{42}
\providecommand{\natexlab}[1]{#1}
\providecommand{\url}[1]{\texttt{#1}}
\expandafter\ifx\csname urlstyle\endcsname\relax
  \providecommand{\doi}[1]{doi: #1}\else
  \providecommand{\doi}{doi: \begingroup \urlstyle{rm}\Url}\fi

\bibitem[Garrett et~al.(2020)Garrett, Lozano-P{\'e}rez, and Kaelbling]{garrett2020pddlstream}
C.~R. Garrett, T.~Lozano-P{\'e}rez, and L.~P. Kaelbling.
\newblock {PDDLStream}: Integrating symbolic planners and blackbox samplers via optimistic adaptive planning.
\newblock In \emph{Proceedings of the International Conference on Automated Planning and Scheduling}, volume~30, pages 440--448, 2020.

\bibitem[Plaku and Hager(2010)]{plaku2010sampling}
E.~Plaku and G.~D. Hager.
\newblock Sampling-based motion and symbolic action planning with geometric and differential constraints.
\newblock In \emph{2010 IEEE International Conference on Robotics and Automation}, pages 5002--5008. IEEE, 2010.

\bibitem[Toussaint(2015)]{toussaint2015logic}
M.~Toussaint.
\newblock Logic-geometric programming: An optimization-based approach to combined task and motion planning.
\newblock In \emph{Twenty-Fourth International Joint Conference on Artificial Intelligence}, 2015.

\bibitem[Kaelbling and Lozano-Pérez(2011)]{kaelbling2011hierarchical}
L.~P. Kaelbling and T.~Lozano-Pérez.
\newblock Hierarchical task and motion planning in the now.
\newblock In \emph{2011 IEEE International Conference on Robotics and Automation}, pages 1470--1477, 2011.

\bibitem[Srivastava et~al.(2014)Srivastava, Fang, Riano, Chitnis, Russell, and Abbeel]{srivastava2014combined}
S.~Srivastava, E.~Fang, L.~Riano, R.~Chitnis, S.~Russell, and P.~Abbeel.
\newblock Combined task and motion planning through an extensible planner-independent interface layer.
\newblock In \emph{2014 IEEE International Conference on Robotics and Automation (ICRA)}, pages 639--646. IEEE, 2014.

\bibitem[Kim et~al.(2019)Kim, Wang, Kaelbling, and Lozano-P{\'e}rez]{kim2019learning}
B.~Kim, Z.~Wang, L.~P. Kaelbling, and T.~Lozano-P{\'e}rez.
\newblock Learning to guide task and motion planning using score-space representation.
\newblock \emph{The International Journal of Robotics Research}, 38\penalty0 (7):\penalty0 793--812, 2019.

\bibitem[Chitnis et~al.(2016)Chitnis, Hadfield-Menell, Gupta, Srivastava, Groshev, Lin, and Abbeel]{chitnis2016guided}
R.~Chitnis, D.~Hadfield-Menell, A.~Gupta, S.~Srivastava, E.~Groshev, C.~Lin, and P.~Abbeel.
\newblock Guided search for task and motion plans using learned heuristics.
\newblock In \emph{2016 IEEE International Conference on Robotics and Automation (ICRA)}, pages 447--454. IEEE, 2016.

\bibitem[Dantam et~al.(2016)Dantam, Kingston, Chaudhuri, and Kavraki]{dantam2016incremental}
N.~T. Dantam, Z.~K. Kingston, S.~Chaudhuri, and L.~E. Kavraki.
\newblock Incremental task and motion planning: A constraint-based approach.
\newblock In \emph{Robotics: Science and systems}, volume~12, page 00052. Ann Arbor, MI, USA, 2016.

\bibitem[Khodeir et~al.(2023)Khodeir, Agro, and Shkurti]{khodeir_ral}
M.~Khodeir, B.~Agro, and F.~Shkurti.
\newblock Learning to search in task and motion planning with streams.
\newblock \emph{IEEE Robotics and Automation Letters}, 8\penalty0 (4):\penalty0 1983--1990, 2023.

\bibitem[Chitnis et~al.(2021)Chitnis, Silver, Kim, Kaelbling, and Lozano-Perez]{Chitnis2020CAMPsLC}
R.~Chitnis, T.~Silver, B.~Kim, L.~Kaelbling, and T.~Lozano-Perez.
\newblock {CAMP}s: Learning context-specific abstractions for efficient planning in factored {MDP}s.
\newblock In \emph{Conference on robot learning}, pages 64--79. PMLR, 2021.

\bibitem[Dhakal et~al.(2023)Dhakal, Talukder, and Stein]{dhakal2023anticipatory}
R.~Dhakal, M.~R.~H. Talukder, and G.~J. Stein.
\newblock Anticipatory planning: Improving long-lived planning by estimating expected cost of future tasks.
\newblock In \emph{International Conference on Robotics and Automation (ICRA)}, pages 11538--11545, 2023.

\bibitem[Arora et~al.(2024)Arora, Singh, Swaminathan, Datta, Banerjee, Bhowmick, Jatavallabhula, Sridharan, and Krishna]{task-anticipation}
R.~Arora, S.~Singh, K.~Swaminathan, A.~Datta, S.~Banerjee, B.~Bhowmick, K.~Jatavallabhula, M.~Sridharan, and M.~Krishna.
\newblock Anticipate \& {Act}: Integrating {LLM}s and classical planning for efficient task execution in household environments.
\newblock In \emph{International Conference on Robotics and Automation (ICRA)}, 2024.

\bibitem[Patel and Chernova(2022)]{patel2022proactive}
M.~Patel and S.~Chernova.
\newblock Proactive robot assistance via spatio-temporal object modeling.
\newblock In \emph{6th Annual Conference on Robot Learning}, 2022.

\bibitem[Garrett et~al.(2021)Garrett, Chitnis, Holladay, Kim, Silver, Kaelbling, and Lozano-P{\'e}rez]{garrett2021integrated}
C.~R. Garrett, R.~Chitnis, R.~Holladay, B.~Kim, T.~Silver, L.~P. Kaelbling, and T.~Lozano-P{\'e}rez.
\newblock Integrated task and motion planning.
\newblock \emph{Annual review of control, robotics, and autonomous systems}, 4:\penalty0 265--293, 2021.

\bibitem[Zhao et~al.(2024)Zhao, Chen, Ding, Zhou, Zhang, Xu, and Zhao]{zhao2024survey}
Z.~Zhao, S.~Chen, Y.~Ding, Z.~Zhou, S.~Zhang, D.~Xu, and Y.~Zhao.
\newblock A survey of optimization-based task and motion planning: From classical to learning approaches.
\newblock \emph{arXiv preprint arXiv:2404.02817}, 2024.

\bibitem[Curtis et~al.(2022)Curtis, Silver, Tenenbaum, Lozano-P{\'e}rez, and Kaelbling]{curtis2022discovering}
A.~Curtis, T.~Silver, J.~B. Tenenbaum, T.~Lozano-P{\'e}rez, and L.~Kaelbling.
\newblock Discovering state and action abstractions for generalized task and motion planning.
\newblock In \emph{Proceedings of the AAAI Conference on Artificial Intelligence}, pages 5377--5384, 2022.

\bibitem[Bradley and Roy(2022)]{bradley2022learning}
C.~P. Bradley and N.~Roy.
\newblock Learning to guide search in long-horizon task and motion planning.
\newblock In \emph{CoRL 2022 Workshop on Learning, Perception, and Abstraction for Long-Horizon Planning}, 2022.

\bibitem[Shah et~al.(2019)Shah, Krasheninnikov, Alexander, Abbeel, and Dragan]{shah2019preferences}
R.~Shah, D.~Krasheninnikov, J.~Alexander, P.~Abbeel, and A.~Dragan.
\newblock The implicit preference information in an initial state.
\newblock In \emph{International Conference on Learning Representations}, 2019.

\bibitem[Ravichandar et~al.(2020)Ravichandar, Polydoros, Chernova, and Billard]{ravichandar2020lfdsurvey}
H.~Ravichandar, A.~S. Polydoros, S.~Chernova, and A.~Billard.
\newblock Recent advances in robot learning from demonstration.
\newblock \emph{Annual Review of Control, Robotics, and Autonomous Systems}, 3:\penalty0 297--330, 2020.

\bibitem[Manschitz et~al.(2014)Manschitz, Kober, Gienger, and Peters]{manschitz2014learning}
S.~Manschitz, J.~Kober, M.~Gienger, and J.~Peters.
\newblock Learning to sequence movement primitives from demonstrations.
\newblock In \emph{International Conference on Intelligent Robots and Systems (IJRR)}, pages 4414--4421, 2014.

\bibitem[Whitney et~al.(2018)Whitney, Rosen, and Tellex]{whitney2018learning}
D.~Whitney, E.~Rosen, and S.~Tellex.
\newblock Learning from crowdsourced virtual reality demonstrations.
\newblock In \emph{International Workshop on Virtual, Augmented, and Mixed Reality for HRI (VAM-HRI)}, 2018.

\bibitem[Moro et~al.(2018)Moro, Nejat, and Mihailidis]{moro2018learning}
C.~Moro, G.~Nejat, and A.~Mihailidis.
\newblock Learning and personalizing socially assistive robot behaviors to aid with activities of daily living.
\newblock \emph{ACM Transactions on Human-Robot Interaction (THRI)}, 7\penalty0 (2):\penalty0 1--25, 2018.

\bibitem[Wang et~al.(2018)Wang, Garrett, Kaelbling, and Lozano-P{\'e}rez]{wang2018active}
Z.~Wang, C.~R. Garrett, L.~P. Kaelbling, and T.~Lozano-P{\'e}rez.
\newblock Active model learning and diverse action sampling for task and motion planning.
\newblock In \emph{2018 IEEE/RSJ International Conference on Intelligent Robots and Systems (IROS)}, pages 4107--4114. IEEE, 2018.

\bibitem[Chitnis et~al.(2019)Chitnis, Kaelbling, and Lozano-P{\'e}rez]{chitnis2019learning}
R.~Chitnis, L.~P. Kaelbling, and T.~Lozano-P{\'e}rez.
\newblock Learning quickly to plan quickly using modular meta-learning.
\newblock In \emph{2019 International Conference on Robotics and Automation (ICRA)}, pages 7865--7871. IEEE, 2019.

\bibitem[Battaglia et~al.(2018)Battaglia, Hamrick, Bapst, Sanchez-Gonzalez, Zambaldi, Malinowski, Tacchetti, Raposo, Santoro, Faulkner, et~al.]{battaglia2018relational}
P.~W. Battaglia, J.~B. Hamrick, V.~Bapst, A.~Sanchez-Gonzalez, V.~Zambaldi, M.~Malinowski, A.~Tacchetti, D.~Raposo, A.~Santoro, R.~Faulkner, et~al.
\newblock Relational inductive biases, deep learning, and graph networks.
\newblock \emph{arXiv preprint arXiv:1806.01261}, 2018.

\bibitem[Shi et~al.(2021)Shi, Huang, shikun feng, Zhong, Wang, and Sun]{Shi2020MaskedLP}
Y.~Shi, Z.~Huang, shikun feng, H.~Zhong, W.~Wang, and Y.~Sun.
\newblock Masked label prediction: Unified message passing model for semi-supervised classification, 2021.

\bibitem[Hamilton et~al.(2017)Hamilton, Ying, and Leskovec]{hamilton2017inductive}
W.~Hamilton, Z.~Ying, and J.~Leskovec.
\newblock Inductive representation learning on large graphs.
\newblock \emph{Advances in neural information processing systems}, 30, 2017.

\bibitem[Silver et~al.(2021{\natexlab{a}})Silver, Chitnis, Curtis, Tenenbaum, Lozano-Perez, and Kaelbling]{silver2021planning}
T.~Silver, R.~Chitnis, A.~Curtis, J.~B. Tenenbaum, T.~Lozano-Perez, and L.~P. Kaelbling.
\newblock Planning with learned object importance in large problem instances using graph neural networks.
\newblock In \emph{Proceedings of the AAAI conference on artificial intelligence}, volume~35, pages 11962--11971, 2021{\natexlab{a}}.

\bibitem[Silver et~al.(2021{\natexlab{b}})Silver, Chitnis, Tenenbaum, Kaelbling, and Lozano-P{\'e}rez]{silver2021learning}
T.~Silver, R.~Chitnis, J.~Tenenbaum, L.~P. Kaelbling, and T.~Lozano-P{\'e}rez.
\newblock Learning symbolic operators for task and motion planning.
\newblock In \emph{2021 IEEE/RSJ International Conference on Intelligent Robots and Systems (IROS)}, pages 3182--3189. IEEE, 2021{\natexlab{b}}.

\bibitem[Kim and Shimanuki(2020)]{kim2020learning}
B.~Kim and L.~Shimanuki.
\newblock Learning value functions with relational state representations for guiding task-and-motion planning.
\newblock In \emph{Conference on Robot Learning}, pages 955--968. PMLR, 2020.

\bibitem[Bertsimas and Tsitsiklis(1993)]{sim_annealing}
D.~Bertsimas and J.~Tsitsiklis.
\newblock {Simulated Annealing}.
\newblock \emph{Statistical Science}, 8\penalty0 (1):\penalty0 10 -- 15, 1993.

\bibitem[Fey and Lenssen(2019)]{pyg}
M.~Fey and J.~E. Lenssen.
\newblock Fast graph representation learning with {PyTorch Geometric}.
\newblock In \emph{ICLR Workshop on Representation Learning on Graphs and Manifolds}, 2019.

\bibitem[Duchi et~al.(2011)Duchi, Hazan, and Singer]{duchi11a}
J.~Duchi, E.~Hazan, and Y.~Singer.
\newblock Adaptive subgradient methods for online learning and stochastic optimization.
\newblock \emph{Journal of Machine Learning Research}, 12\penalty0 (61):\penalty0 2121--2159, 2011.

\bibitem[Coumans and Bai(2016--2021)]{coumans2021}
E.~Coumans and Y.~Bai.
\newblock {PyBullet}, a {Python} module for physics simulation for games, robotics and machine learning, 2016--2021.

\bibitem[Liu et~al.(2022)Liu, Liu, Qin, Xiang, Gou, Xin, Roa, Calli, Su, Sun, and Tan]{liu2021ocrtoc}
Z.~Liu, W.~Liu, Y.~Qin, F.~Xiang, M.~Gou, S.~Xin, M.~A. Roa, B.~Calli, H.~Su, Y.~Sun, and P.~Tan.
\newblock {OCRTOC}: A cloud-based competition and benchmark for robotic grasping and manipulation.
\newblock \emph{IEEE Robotics and Automation Letters}, 7\penalty0 (1):\penalty0 486--493, 2022.

\bibitem[{Fetch Robotics}(2023)]{fetch2023}
{Fetch Robotics}.
\newblock The {Fetch Mobile Manipulator}.
\newblock \url{https://fetchrobotics.borealtech.com/robotics-platforms/fetch-mobile-manipulator/?lang=en}, 2023.

\bibitem[Fox and Long(2003)]{fox2003pddl2}
M.~Fox and D.~Long.
\newblock {PDDL2.1}: An extension to {PDDL} for expressing temporal planning domains.
\newblock \emph{Journal of Artificial Intelligence Research}, 20:\penalty0 61--124, 2003.

\bibitem[Helmert(2006)]{helmert-jair2006}
M.~Helmert.
\newblock The {Fast} {Downward} planning system.
\newblock \emph{Journal of Artificial Intelligence Research}, 26:\penalty0 191--246, 2006.

\bibitem[Kuffner and LaValle(2000)]{Kuffner2000RRTconnectAE}
J.~J. Kuffner and S.~M. LaValle.
\newblock Rrt-connect: An efficient approach to single-query path planning.
\newblock \emph{Proceedings 2000 ICRA. Millennium Conference. IEEE International Conference on Robotics and Automation. Symposia Proceedings (Cat. No.00CH37065)}, 2:\penalty0 995--1001 vol.2, 2000.

\bibitem[Quigley et~al.(2009)Quigley, Conley, Gerkey, Faust, Foote, Leibs, Wheeler, Ng, et~al.]{quigley2009ros}
M.~Quigley, K.~Conley, B.~Gerkey, J.~Faust, T.~Foote, J.~Leibs, R.~Wheeler, A.~Y. Ng, et~al.
\newblock {ROS}: an open-source robot operating system.
\newblock In \emph{ICRA workshop on open source software}, volume~3, page~5, 2009.

\bibitem[Olson(2011)]{apriltag}
E.~Olson.
\newblock {AprilTag}: A robust and flexible visual fiducial system.
\newblock In \emph{2011 IEEE International Conference on Robotics and Automation}, pages 3400--3407, 2011.

\bibitem[Coleman et~al.(2014)Coleman, Sucan, Chitta, and Correll]{coleman2014moveit}
D.~T. Coleman, I.~A. Sucan, S.~Chitta, and N.~Correll.
\newblock Reducing the barrier to entry of complex robotic software: a {MoveIt!} case study.
\newblock \emph{Journal of Software Engineering in Robotics}, 5\penalty0 (1):\penalty0 3--16, 2014.

\end{thebibliography}

\clearpage{}
\appendix
\section{Appendix}

\lstdefinelanguage{PDDL}
{
  sensitive=false,    
  morecomment=[l]{;}, 
  alsoletter={:,-},   
  morekeywords={
    define,domain,problem,not,and,or,when,forall,exists,either,
    :domain,:requirements,:types,:objects,:constants,
    :predicates,:action,:parameters,:precondition,:effect,
    :fluents,:primary-effect,:side-effect,:init,:goal,
    :strips,:adl,:equality,:typing,:conditional-effects,
    :negative-preconditions,:disjunctive-preconditions,
    :existential-preconditions,:universal-preconditions,:quantified-preconditions,
    :functions,assign,increase,decrease,scale-up,scale-down,
    :metric,minimize,maximize,
    :durative-actions,:duration-inequalities,:continuous-effects,
    :durative-action,:duration,:condition
  }
}
\lstset{
    basicstyle=\ttfamily\scriptsize,
    keywordstyle=\color[rgb]{0.3,0.1,0.5},
    linewidth=\columnwidth
}

\subsection{Task and Motion Planning Overview and Solver Details}

The \glsentryfull{TAMP} solver we leverage for this work relies on the planning strategy presented in the work of \citet{srivastava2014combined}.
It operates by first generating a high-level symbolic \emph{task skeleton}: a sequence of high-level symbolic actions.
However, a symbolic plan alone is not always actionable, as it requires further refinement at the motion level.
Thus, the task skeleton must be \emph{refined}, a process by which the abstract actions in the skeleton are assigned metric parameters: e.g., how to grasp an object or where in the cabinet a bowl should be placed. 
The resulting metric plans are checked for feasibility by a low-level motion planner, checking for instance if a collision-free trajectory can be found that achieves the desired motion between two sampled states. If the plan is determined infeasible, new metric parameters are sampled, yielding a new metric plan. After sufficiently many failed attempts, a new task skeleton may be generated. This approach of searching over both the high-level symbolic plan and the low-level metric plan continues until a feasible solution to the assigned task is found.

For the \textsc{TAMPSolver} used in Algorithm~\ref{alg:anttamp}, the \glsentryfull{PDDL}~\cite{fox2003pddl2} is used to define the operators (action schemas) available to the robot. The popular FastDownward solver~\cite{helmert-jair2006} is used to generate the high-level task skeletons, which are then refined iteratively by a sampler and, when necessary, a motion-level planner.
Motion planning---e.g., for the robot's movement between poses---is determined via the RRT-Connect algorithm~\cite{Kuffner2000RRTconnectAE}, yielding costs based on Euclidean distances that are then fed back into the planner.

\subsection{PDDL Example Code Listing}

Our experiments make use of planning environments that appear in the work of \citet{Chitnis2020CAMPsLC}, adapted to be useful for our sequential \gls{TAMP} setting.
Here, we show an example \gls{PDDL} operator definition from our cabinet-loading domain of Sec.~\ref{sec:exp:cabinet}:

\begin{lstlisting} [language=PDDL, linewidth=\columnwidth, caption=PDDL operator definition for \texttt{pick} in Cabinet domain, label={lst:cabinetpddl}]
(:action pick
 :parameters (?o - object ?grasppose - pose ?pickpose - pose)
 :precondition (and (at ?o ?pickpose)
                    (isgrasppose ?graspose ?o ?pickpose)
                    (forall (?o2 - object) (not (objobstructs ?o2 ?grasppose)))
                    (emptygripper)
                    (not (clear ?pickpose)))
 :effect (and (not (at ?o pickpose)) (not (emptygripper))
              (ingripper ?o) (clear ?pickpose)
              (forall (?p2 - pose) (not (objobstructs ?o ?p2)))
              (increase (cost) 20))
)
\end{lstlisting}
Notably, the pose parameters are sampled during plan refinement and the predicate function \texttt{objobstructs} specifies the feasibility of a chosen grasp, determined via a separate collision checking process using the motion planner.
Our cabinet domain features three operators: \texttt{move}, \texttt{pick}, and \texttt{place}; the costs of the \texttt{pick} and \texttt{place} operators have a fixed 20 units of cost and the \texttt{move} operator has a variable cost depending on the Euclidean distance of movement. The \gls{NAMO} environment of Sec.~\ref{sec:exp:namo} features a single \texttt{moveclear} operator, which moves to a block, moving other block out of the way in its effort to reach it. It has a base cost of 200 units per block moved and an additional cost corresponding to the Euclidean distance of the robot's total movement.

\clearpage{}
\subsection{Preparation for Anticipatory \Gls{TAMP}}
\begin{wrapfigure}{R}{0.48\textwidth}
\vspace{-2.5em}
\begin{minipage}{0.48\textwidth}
\begin{algorithm}[H]
\caption{Preparation}\label{alg:preptamp}
\algrenewcommand{\algorithmiccomment}[1]{\hfill\textbf{//}\,#1}
\scriptsize
\begin{algorithmic}[0]
\Function{Preparation}{$s_0$, \textsc{APCostEstimator}}
    \LineComment{Initialize temperature and cooling rate}
    \State temperature = 1000; cooling\_rate = 0.95
    \LineComment{Estimate the $\apcost{}$ for initial state $s_0$.}
    \State $\apcost{}(s_0) =\textsc{APCostEstimator}(s_0)$
    \LineComment{Initialize initial state to current state and prepared state.}
    \State $s_{\text{prep}} = s_{\text{current}} = s_0$
    \State $\apcost{}(s_{\text{current}}) = \apcost{}(s_0)$
    \For{$i \in \{1, 2, \dots, N\}$}
        \LineComment{Sample new random state.}
        \State $s_{\text{new}}$= \textsc{GetNeighborState}($s_{\text{current}}$)
        \LineComment{Estimate the $\apcost{}$ for the new random state.}
        \State $\apcost{}(s_{\text{new}})$ = \textsc{APCostEstimator}($s_{\text{new}}$)
        \LineComment{Determine whether to accept the new state.}
        \State $\delta =  \apcost{}(s_{\text{new}})-\apcost{}(s_{\text{current}})$ 
        \If{$\delta < 0$ \textbf{or} random() $<$ exp($-\delta $ / temperature)}
            \State $s_{\text{current}} = s_{\text{new}}$
            \State $\apcost{}(s_{\text{current}}) = \apcost{}(s_{\text{new}})$
            \If{$\apcost{}(s_{\text{current}}) < \apcost{}(s_{\text{prep}})$}
                \State $s_{\text{prep}} = s_{\text{current}}$
                \State $\apcost{}(s_{\text{prep}}) = \apcost{}(s_{\text{current}})$
            \EndIf
        \EndIf
        \State temperature = temperature $\times$ cooling\_rate
    \EndFor
    \State \Return $s_{\text{prep}}$
\EndFunction
\end{algorithmic}
\end{algorithm}
\end{minipage}
\end{wrapfigure}
Task-free anticipatory \gls{TAMP} or \emph{preparation} involves minimizing the anticipatory planning cost in advance of receiving any tasks. We use simulated annealing~\cite{sim_annealing} for the optimization process, as shown in Algorithm~\ref{alg:preptamp}. Preparation, defined in Eq.~\eqref{eq:prep}, involves searching over the continuous states of the environment an environment to find the state that minimizes the anticipatory planning cost (the expected future cost) as estimated by the $\textsc{APCostEstimator}$. The algorithm starts from the initial state of the environment and explores neighboring states by perturbing object positions using the $\textsc{GetNeighborState}$ function.
New states are accepted---i.e., used for the subsequent iteration---if the new state improves upon the expected future cost or is randomly accepted via a monotonically-decreasing probabilistic acceptance criteria, which exists to avoid settling in local minima.
The process iterates for a set number of times, with a \emph{cooling schedule} gradually reducing the probabilistic acceptance criteria, and thus the likelihood of uphill moves. Finally, the state with the lowest cost is returned. It is this state into which the robot puts the environment and from which planning will then commence.


\subsection{Real Robot Experiment Details}
\begin{wrapfigure}{R}{0.5\textwidth}
    \vspace{-1em}
    \includegraphics[width=0.5\textwidth]{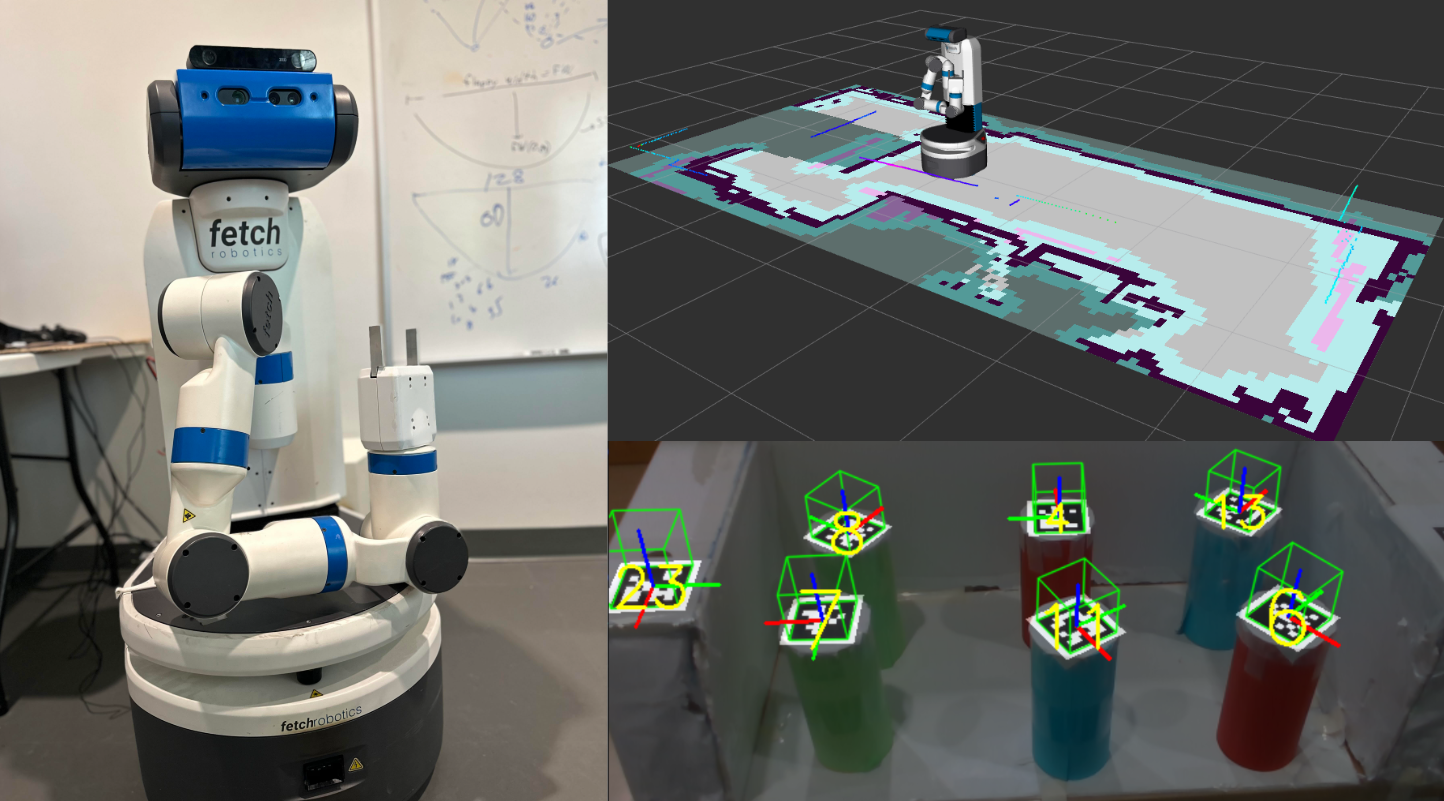}
    \caption{Left: Our Fetch robot. Top right : Mapping of the environment shown in Rviz. Bottom right: Use of AprilTags for perception.}\label{fig:fetch_map}
    \vspace{-0.5em}
\end{wrapfigure}
We use a Fetch Mobile Manipulator (Figure~\ref{fig:fetch_map}) for real world demonstration. Our Fetch robot primarily relies on three sub-processes: navigation, perception, and manipulation. In the navigation process, the Fetch uses a pre-built map of the environment as shown at the top right of Figure~\ref{fig:fetch_map} and the ROS move\_base package~\cite{quigley2009ros} to localize itself within the map using its laser scanner and move between specified locations. For perception, we utilize AprilTag~\cite{apriltag} fiducial markers to estimate the poses of targeted objects relative to the robot, as shown at the bottom right of Figure~\ref{fig:fetch_map}. These poses are transformed to the robot frame and published as a separate ROS node, to which other processes subscribe for pose data. For manipulation, the MoveIt!~package~\cite{coleman2014moveit} is used to determine feasible grasp poses and trajectories for the robot's arm. The outputs of the perception node are used to create a 3D representation of the environment, which the IKFast solver~\cite{quigley2009ros, coleman2014moveit} uses to calculate the arm configuration that will achieve the desired end effector pose. The MoveIt!~motion planner then generates the trajectory for the arm to pick or place objects in locations specified by the planner.
The demonstrations involve first generating a plan in the simulated environments using the \gls{TAMP} solver described above and then execute that plan, by sequential execution of the actions it prescribes, using the aforementioned planning and perceptual modules. Videos of the demonstrations are included at the end of the video presentation included in the supplementary material.


\subsection{Example 10-Task Sequence in \textsc{Namo} domain}
\begin{figure}[H]
    \centering
    \includegraphics[width=\textwidth]{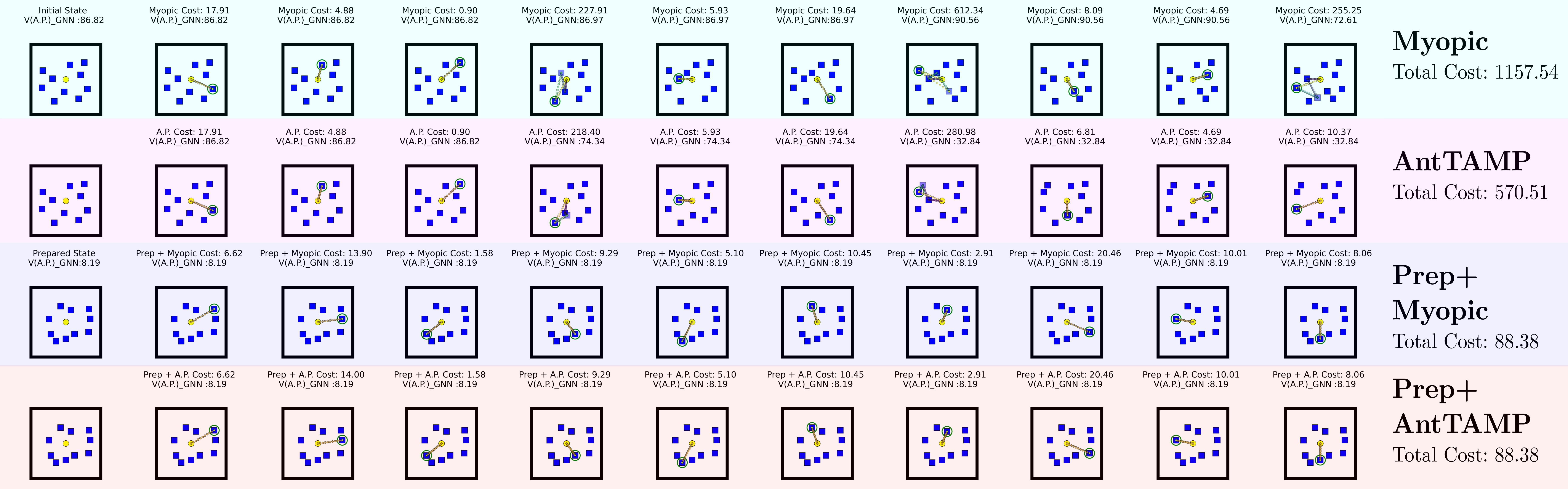}
    \caption{An experiment trial in a \gls{NAMO} for a 10-task sequence. Above each state, we show both immediate planning costs for each completed task and $\apcost{}$ as estimated by the \gls{GNN}.}\label{fig:namo_longer}
\end{figure}

We present an example trial in the \gls{NAMO} domain involving a sequence of 10 tasks in a persistent environment, each specifying a particular object the robot must reach (see Figure~\ref{fig:namo_longer}). Our \anttamp{} approach reduces the overall planning time by 50\%. Supporting our results from Sec.~\ref{sec:exp:namo}, preparing the environment in advance results in a configuration from which all blocks can be easily reached without moving any others out of the way, an emergent behavior of our approach. As such, the behavior of both \myopic{} and our \anttamp{} result in a 92\% reduction in overall planning cost compared to \myopic{} planning in the unprepared environment.

\subsection{Example 10-Task Sequence in Cabinet-Loading Domain}
\begin{figure}[H]
    \centering
    \includegraphics[width=\textwidth]{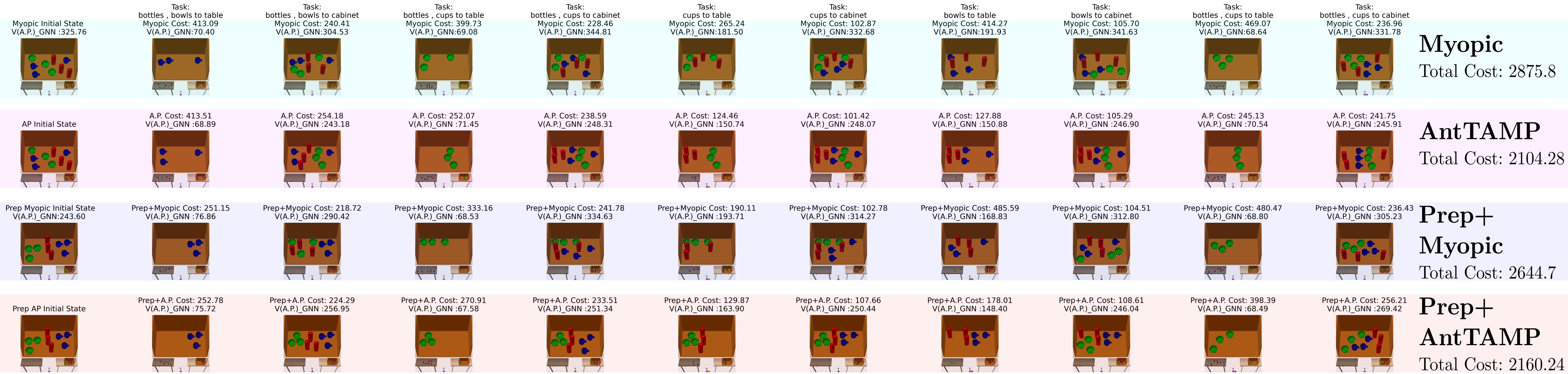}
    \caption{An example trial in a Cabinet-loading domain for a 10-task sequence. Above each state, we show both immediate planning costs for each completed task and $\apcost{}$ as estimated by the \gls{GNN}.}\label{fig:cabinet_longer}
\end{figure}

We also present an example 10-task sequence in our cabinet-loading domain, as shown in Figure~\ref{fig:cabinet_longer}. Consistent with the results shown in Sec.~\ref{sec:exp:cabinet}, our \anttamp{} approach reduces planning costs throughout the sequence, particularly during the unloading of objects from the cabinet.  Using our approach, the overall planning cost of the sequence is reduced by 26\% compared to \myopic{}. We additionally show how preparation can further reduce cost of myopic planning, though the benefit of initially preparing the state diminishes as the task sequence proceeds. In this particular trial, both the prepared and non-prepared anticipatory \gls{TAMP} results are similar, with only a small difference between them. Our results show how our approach improves planning cost even over a longer sequence of tasks.

\end{document}